%% file: ijcai22.tex
\documentclass{article}
\usepackage{ijcai22}

\usepackage{times}
\usepackage{soul}
\usepackage{url}
\usepackage{hyperref}
\hypersetup{colorlinks,allcolors=black}
\usepackage[utf8]{inputenc}
\usepackage[small]{caption}
\usepackage{graphicx}
\usepackage{amsmath}
\usepackage{amsthm}
\usepackage{booktabs}
\usepackage{algorithm}
\usepackage{algorithmic}
\usepackage{enumitem}
\usepackage{amsfonts} 
\newcommand{\ie}[0]{\textit{i.e.}, }

\usepackage{float}
\usepackage{placeins} 
\usepackage{comment}

\title{Fast Server Learning Rate Tuning for Coded Federated Dropout}

\author{
Giacomo Verardo,
Daniel Barreira,
Marco Chiesa,
Dejan Kostic,
Gerald Q. Maguire Jr.
\affiliations
KTH Royal Institute of Technology 
\emails
\{verardo, barreira, mchiesa, dmk, maguire\}@kth.se
}

\usepackage{subfigure}
\begin{document}

\maketitle

\begin{abstract}

\input{Abstract}
\end{abstract}

\noindent
\input{Introduction}
\input{Motivation}
\input{Methodology}

\input{Evaluation}
\input{Conclusion}

\FloatBarrier

\newpage
\bibliographystyle{named}
\bibliography{ijcai22}
\include{Appendix}

\end{document}

%% file: Abstract.tex
In cross-device Federated Learning (FL), clients with low computational power train a common\linebreak[4] machine model by exchanging parameters via updates instead of potentially private data. Federated Dropout (FD) is a technique that improves the communication efficiency of a FL session by selecting a \emph{subset} of model parameters to be updated in each training round. However, compared to standard FL, FD produces considerably lower accuracy and faces a longer convergence time. In this paper, we leverage \textit{coding theory} to enhance FD by allowing different sub-models to be used at each client. We also show that by carefully tuning the server learning rate hyper-parameter,  we can achieve higher training speed while also achieving up to the same final accuracy as the no dropout case. For the EMNIST dataset, our mechanism achieves 99.6\% of the final accuracy of the no dropout case while requiring $2.43\times$ less bandwidth to achieve this level of accuracy. 



%% file: Introduction.tex
\section{Introduction}

In cross-device Federated Learning (FL)~\cite{mcmahan2017communicationefficient}, the parameter server broadcasts a global machine learning (ML) model to the devices (clients), which in turn perform training over their own datasets. The resulting model updates are sent from the clients to the server, which aggregates the results and may start another FL round. Even in the case of highly heterogeneous client datasets, it has been demonstrated that the model converges~\cite{li2020convergence}. As the size of a model could be hundreds of MB~\cite{kerasModels}, Federated Dropout (FD) has been used to both reduce the size of the model and the size of the updates, reducing both communication costs and computational costs at the clients.

Splitting a common, global model between clients and training it collaboratively has become imperative to reduce both memory and computational demands of an FL session. In FD, a global model is divided into sub-models which are trained locally and then merged into an updated global model.  FD is orthogonal to message compression techniques, such as quantization~\cite{alistarh2017qsgd} or sparsification~\cite{alistarh2018convergenceSParse}, which do not reduce the computational power and memory needed at the client side. 

Reaping the benefits of FD is not easy as it entails solving two main challenges:
\begin{itemize}[leftmargin=*,noitemsep]
    \item \textbf{Low accuracy.} FD learning may result in \textit{lower accuracy} than traditional FL~\cite{caldasMCmahanReducingCommOverhead}. Intuitively, partially overlapping sub-models per client may improve performance. However, it is not clear how sub-models should be selected, nor which is the orthogonality level which produce the optimal accuracy. Moreover, merging the sub-models into a new global model is a challenging task since it depends on their overlapping.
    \item \textbf{Slow convergence.} Although FD requires less bandwidth per round compared to traditional (no-dropout) FL, there is no guarantee that FD will converge rapidly. If FD requires significantly more rounds than FL to achieve high accuracy, then the promised bandwidth benefits would vanish.
\end{itemize}

We propose novel techniques to improve the accuracy of FD \textit{without} losing the inherent bandwidth savings offered by FD (\ie by improving convergence speed). To tackle the above challenges, we explore the following two ideas:
\begin{itemize}[leftmargin=*,noitemsep]
 \item \textbf{\textit{Applying coding theory for sub-model selection.}} Existing results show that sending random sub-models rather than the same model to each client performs better~\cite{wen2021federateddropout}. Building upon the idea of sending different models to different clients during one FD round, we deterministically compute sub-models and then examine whether they perform better than random sub-models. We draw inspirations from \textit{coding theory} (specifically, from the Code Division Multiple Access (CDMA) problem where orthogonal codes enable simultaneous communication channels \emph{without} interference). We employ Gold codes\,\cite{goldcodes} and Constant Weight Codes (CWC)\,\cite{cwcnewtable} as masks to drop units and create different sub-models. The intuition is that selecting sub-models using these mechanisms will produce higher accuracy than random selection. 
 \item \textbf{\textit{Adaptive server learning rate}}. We experimentally observe that the convergence speed of an FD session depends on a critical parameter called \textit{server learning rate}, which determines how the weight updates from the clients are incorporated into the trained model. Thanks to the inherent bandwidth savings of FD, we propose to search for the best server learning rate at the beginning of an FD session. 
 
\end{itemize}
Based on the above ideas, we design a mechanism called Coded Federated Dropout (CFD), which we incorporate alongside existing state-of-the-art FL systems, such as FedAdam~\cite{reddi2020adaptive} and FedAvg~\cite{mcmahan2017communicationefficient}. Based on an evaluation with the EMNIST62 dataset, we show that CFD increases the final accuracy of the trained models while preserving the bandwidth savings of FD.

In summary, our contributions are:
\begin{itemize}[leftmargin=*,noitemsep]
\item We are the first to leverage coding theory to carefully select the sub-models used in each FL round. 
\item We show that the optimal server learning rate in a traditional FL session differs from that of an FD session. 
\item We design a technique to \textit{quickly} search for the best server learning rate. Our evaluation shows that we can identify good server learning rates in just hundreds of rounds.
\item We show that CFD with Gold Codes achieves \textit{comparable accuracy} to no-dropout FL with $2.43\times$ \textit{less bandwidth} on the EMNIST dataset. 
\item We show that minimizing the ``cross-correlation'' metric in Gold Codes produces better final accuracy than maximizing the ``minimal distance'' metric of CWC codes.

\end{itemize}

%% file: Motivation.tex
\section{Background}
\textbf{Federated Learning} An FL session is composed of one parameter server and multiple clients. At the beginning of an FL round, the server broadcasts a common global model $w^{(t)}_{k}$ to a fraction of the clients. At round $t$, each client $k$ trains for a customizable number of epochs E and returns the update $\Delta w^{(t)}_{k}$ from the previously received weights:
\begin{equation}
    \hat{w}^{(t)}_{k} = w^{(t)}_{k}-\eta_{l} \nabla_w L(w,D_k)
\end{equation}
\begin{equation}
    \Delta w^{(t)}_{k} = \hat{w}^{(t)}_{k} - w^{(t)}_{k}
\end{equation}
\noindent where $L(w,D_k)$ is the loss function, which is a function of the model weights and the client dataset $D_k$. 
When employing Federated Averaging (FAVG, \cite{mcmahan2017communicationefficient}), the originally proposed aggregation method, the parameter server computes new weights by averaging the updates and adding them to the previous global model:
\begin{equation}
\label{eq:aggregation}
    w^{(t+1)} = w^{(t)} + \eta \sum_{j \in S(t)}p_j \Delta w^{(t)}_{j}
\end{equation}
\begin{equation}
    p_j = \frac{|D_j|}{\sum_{j \in S(t)}|D_j|}
\end{equation}
 For FAVG, $\eta$ is set to $1.0$, whereas it can be different for other aggregation mechanisms.
 
 \noindent\textbf{Federated Dropout} One of the major issues of FL is the communication overhead. FD improves bandwidth efficiency by randomly dropping connections between adjacent neural network layers. Unlike standard dropout \cite{dropout}, FD is not employed as a regularisation tool. Instead, FD keeps a fixed percentage $\alpha$ of activations, thus producing a sub-model with a $(1-\alpha)^2$ parameters fraction of fully connected networks. The sub-model is trained at the client-side, while the aggregation procedure only involves those nodes that have been kept. Moreover, the required computational power and memory at the client are reduced. However, although the benefits in bandwidth efficiency are remarkable, the same set of weights is trained at each client per round.

 \cite{bouacida2020adaptive} suggest adapting the selection of dropped nodes based on the loss for each client; however, they state that this approach is unsuitable for FL since it wastes too much memory at the server. Therefore, they propose an alternative technique which employs a single sub-model for all clients, which unfortunately degrades convergence rate and final accuracy.
 In contrast to random dropout\,\cite{wen2021federateddropout}, we propose the use of coding theory as a tool for mask selection to enhance sub-models' orthogonality by producing different dropping masks for each client. Hence, we allow partially disjoint sub-models per clients per round to improve both convergence time and final accuracy for the same $\alpha$.

 \noindent\textbf{Code Division Multiple Access} Multiple access techniques address the problem of having multiple users communicate via a shared channel. Time and frequency division multiple access respectively splits the channel in time and frequency between users. In contrast, CDMA assigns a different code to each user and allows each of them to use the whole channel. If the codes are sufficiently orthogonal, the inter-user interference is low and transmission occurs with negligible error rate. CDMA has been extensively used in satellite \cite{cdmasatellite} and mobile communication \cite{cdmamobile},
 . Gold \cite{goldcodes} and Kasami \cite{kasami1966weight} codes are examples of families of sequences designed for orthogonality.
  
  

 \noindent\textbf{Adaptive Federated Optimization} \cite{reddi2020adaptive} have proven that using a different learning rate for each parameter during aggregation can greatly improve FL model convergence. They propose 3 aggregation methods (FedAdam, FedAdagrad, and FedYogi), which replace Eq.\ref{eq:aggregation}. Here we describe only the FedAdam algorithm, which performs well in all the datasets:
  




  \begin{equation}
  \label{eq:fedadam1}
      \Delta^{t} = \beta_{1}\cdot\Delta^{t-1}+(1-\beta_{1})\cdot\sum_{j \in S(t)}p_j \Delta w^{(t)}_{j}
  \end{equation}
  \begin{equation}
  \label{eq:fedadam2}
      v^{(t)} =\beta_{2}\cdot v^{(t-1)} + (1-\beta_{2})\cdot {\Delta^{(t)}}^{2}
  \end{equation}
  \begin{equation}
  \label{eq:fedadam3}
     w^{(t+1)} = w^{(t)} + \eta \frac{\Delta^{(t)}}{\sqrt{v^{(t)}}+\tau}
  \end{equation}
where $\beta_{1}$,$\beta_{2}$ and $\tau$ are hyper-parameters.
The key insight is that training variables which have been trained less in the previous rounds will improve convergence. For this reason, $v_t$ stores an indication of how much variables have been trained and is used to independently scale each component of the next update $\Delta^{(t)}$. However, the proposed optimization techniques require expensive hyper-parameters tuning and therefore a considerable amount of time.

 Other adaptive mechanisms have been proposed to correct the client drift due to the statistical heterogeneity in the clients datasets. \cite{karimireddy2021scaffold} employ variance reduction, but this requires too much information to be stored server-side. \cite{ditto} propose a trade-off between fairness and robustness of the global model. 

%% file: Methodology.tex
\section{Methodology}
This section describes the proposed Coded Federated Dropout (CFD) method which performs both tuning of the server learning rate $\eta$ (Sect.~\ref{sub:adapt}) and the selection of the sub-models sent to the clients (Sect.~\ref{sub:cfd}). 

\subsection{Fast server learning rate adaptation}
\label{sub:adapt}

\begin{figure}[hbt]
\centering
\hspace{-4ex}
\subfigure[No Dropout]{
\includegraphics[height=1.4in]{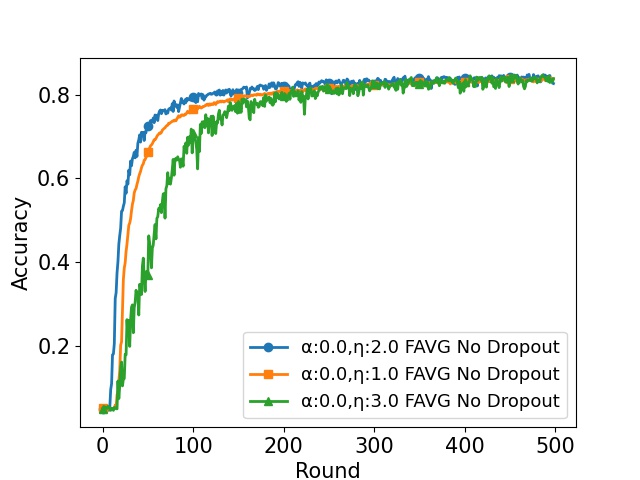}\label{fig:motivation-a}}
\hspace{-4.8ex}
\subfigure[Federated Dropout]{
\includegraphics[height=1.4in]{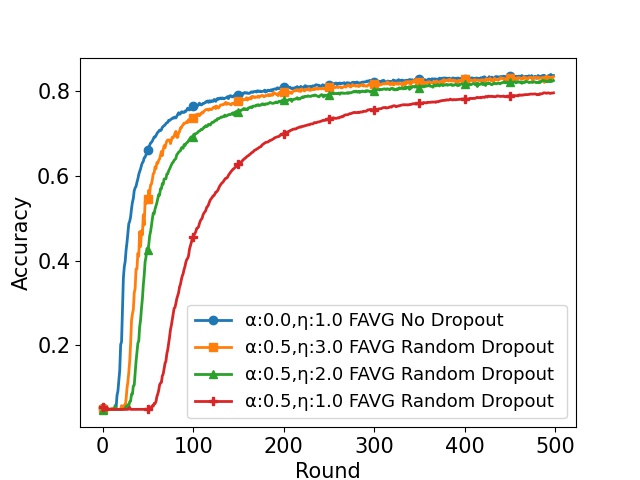}\label{fig:motivation-b}}
\caption{Increasing the server learning rate $\eta$ to 3.0 is beneficial for random FD with different sub-models per client, while it is detrimental for the no dropout case}
\label{fig:motivation}
\end{figure}

Similarly to centralized ML, increasing the server learning rate may lead to faster convergence, but further increasing the learning rate causes the objective function to diverge\,\cite{zeiler2012adadelta}. Fig.~\ref{fig:motivation-a} empirically confirms that this is also the case for no-dropout FL where a high server learning rate of $\eta = 3$ exhibits worse convergence than with $\eta=2$. This result is based on the EMNIST62 dataset with more details in Sect.~4. Interestingly, Fig.~\ref{fig:motivation-b} shows that in FD with random sub-models increasing the server learning rate to $3$ leads to faster convergence. This shows that the ``best'' server learning rate for FD may differ from the no-dropout case. 

We propose a fast server learning rate adaptation method, which can also be extended to other parameters. At the beginning of training, we run Algorithm~\ref{alg:fslra}, which requires $n_a$ adaptation steps. In each step (line 2), multiple FL sessions are launched in parallel from the same parameter server with different server learning rates $\mathbb{H}$ and, in general, different clients subsets per round. We start our search using three $\eta$ values during the first adaptation step (line 1) and reduce it to two server learning rates in the following adaptation steps (line 14). The goal of this search (lines 4--12) is to find the server learning rate that reaches a preconfigured accuracy target $\gamma$ (lines 8--9) in the minimum number of rounds $r^*$ (line 5). More specifically, in each round $r$, the server collects both the gradient update $\Delta w^{t}_k$ and the accuracy of the model for each training client in the round (line 6). Then, it computes the average of the median training accuracy $\overline{\gamma}$ in the last $q$ FL rounds (line 7). The median operation is performed in order to avoid the impact of outliers (\ie clients with too high or low training accuracies), when the average over the last rounds avoids sudden spikes. If one server learning rate $\overline{\gamma}$ is higher than a predefined threshold $\gamma^{*}$, then we have found a new optimal server learning rate $\eta^{*} = \eta$ that requires the new minimum number of rounds $r^{*} = r$ to achieve the target accuracy (lines 8--9). For the next adaptation step, the new tentative server learning rates $\mathbb{H}$ are chosen near $\eta^{*}$ (lines 15--16) and the next adaptation step is performed. An adaptation step may also end when all the FL sessions produce $r >= r^{*}$ (line 5). Worth noting is that the search at lines 3--13 can be done in parallel to improve convergence speed.
This algorithm reduces the number of rounds compared to testing all possible server learning rates using full FL sessions. In particular, since sessions are aborted when $r >= r^{*}$, the overhead introduced by each adaptation step is limited. Assuming the parallel search is synchronized round-by-round, the additional overhead in number of rounds of our algorithm is: 
\begin{equation}
\label{eq:commoverhead}
    3\cdot r^*_0 + 2\sum_{i=1}^{n_a}r^*_i - r^*
\end{equation}

\noindent
where $r^*_i$ is the minimum number of rounds at the end of the adaptive step $i$. The first term accounts for the first adaptive step, the second terms for the following adaptive steps, and the third term for the spared training rounds when running the full simulation with $\eta=\eta^{*}$.

Algorithm~\ref{alg:fslra} selects the optimal $\eta^{*}$ in terms of number of rounds to reach the target accuracy. The Round function at line 6 represents the underlying FL mechanism being used to compute the trained model, for instance, FAVG or FedAdam.

We argue that running full simulations to achieve the same level of granularity takes $T\times$ the number of rounds per simulation, where T is the number of tried server learning rates. This value is strictly greater than the bound provided by equation \ref{eq:commoverhead}.
\begin{algorithm}[hbt]
\caption{Fast server learning rate adaptation}
\label{alg:fslra}
\textbf{Input}: $w^{0}$, $\{D_k \forall k \in \{1,..., T\}\}$\\
\textbf{Parameter}: $\gamma^{*}$, $q$, $n_a$, $\eta_0$,$\Delta\eta$\\
\textbf{Output}: $\eta^{*}$
\begin{algorithmic}[1]
\STATE $ \mathbb{H} \leftarrow \{ \eta_0 ,\, \eta_0 - \Delta\eta ,\, \eta_0 + \Delta\eta \}$
\FOR{$s=0$ \TO $n_a$}
\FOR{$\eta \in \mathbb{H}$ \textbf{parallel}} 
\STATE $ r\leftarrow 0$
\WHILE{$r<r^{*}$}
\STATE $\overline{\gamma}^{t}, \, w^{(t+1)} \leftarrow Round(\{D_k \} , w^{(t)})$
\STATE $\overline{\gamma} \leftarrow \frac{1}{q} \cdot \sum_{i=0}^{q-1}{\overline{\gamma}^{t-i}}$
\IF{$\overline{\gamma} \geq \gamma^{*}$}
\STATE $ r^{*}\leftarrow r,\, \eta^{*}\leftarrow \eta $
\STATE wait for parallel search to end; go to line 15
\ENDIF
\STATE $ r\leftarrow r+1$
\ENDWHILE
\ENDFOR
\STATE $\Delta\eta \leftarrow \frac{\Delta\eta}{2}$
\STATE $ \mathbb{H} \leftarrow \{ \eta^{*} - \Delta\eta ,\, \eta^{*} + \Delta\eta \}$
\ENDFOR
\end{algorithmic}
\end{algorithm}
\FloatBarrier

\subsection{Coded Federated Dropout}
\label{sub:cfd}
\begin{figure}[!ht]
\centerline{\includegraphics[scale=0.4]{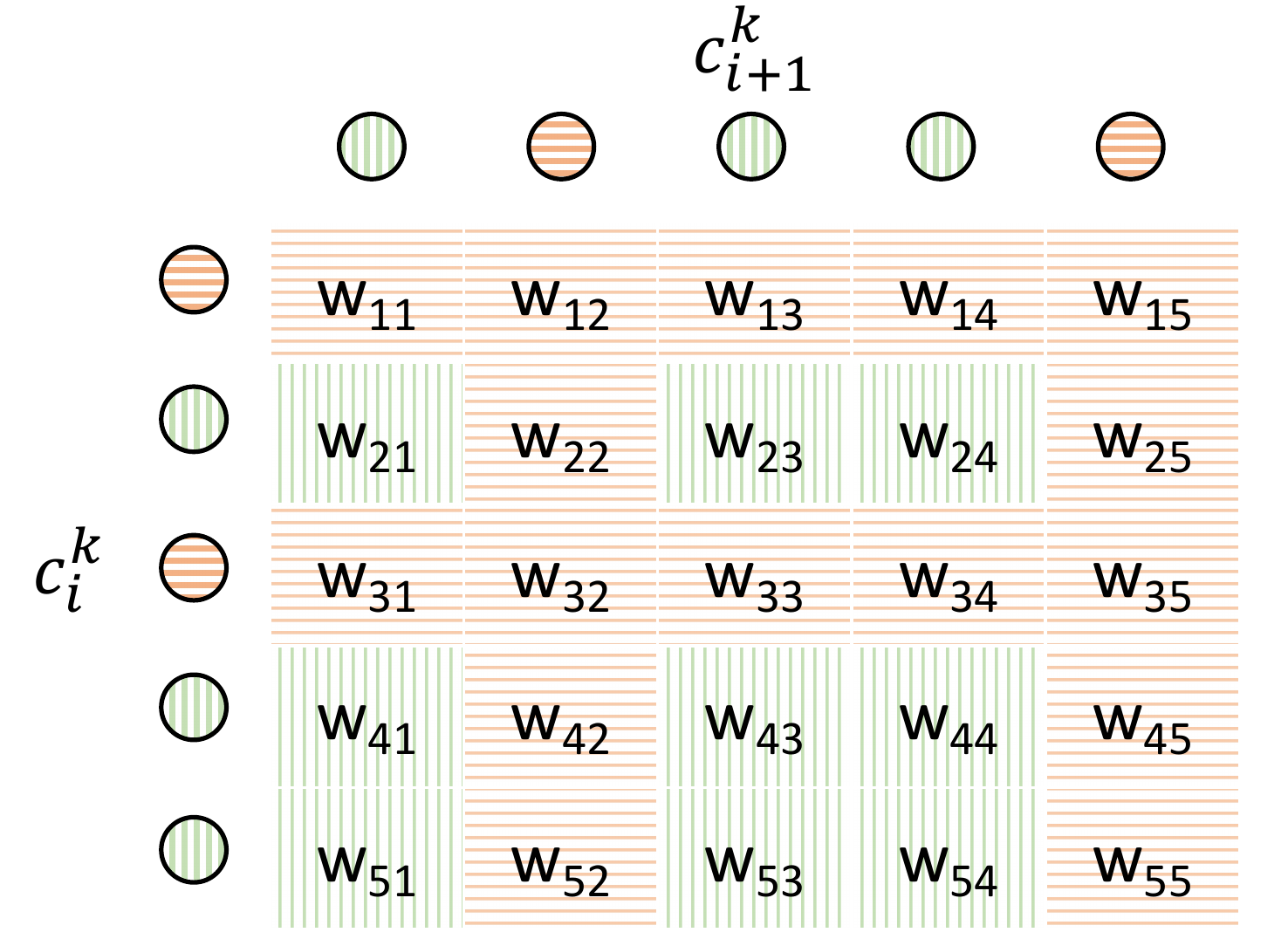}}
\caption{Federated Dropout for masks $c^{k}_{i}$=[0,1,0,1,1] and $c^{k}_{i+1}$=[1,0,1,1,0] reduces the weights to be trained from 25 to 9}
\label{fig:feddropoutexample}
\end{figure}

We reduce the size of the model by dropping weights from each layer by associating to each client $k$ and model layer $i$ in the FL round a binary mask vector $c^{k}_{i}\in \mathbf{R}^{N_{i}}$. A unit is dropped or kept when the component $c^{k}_{ij}$ is equal to 0 or 1 respectively. For adjacent fully connected layers (Fig.~\ref{fig:feddropoutexample}) the dropped weights can be straightforwardly obtained by eliminating rows and columns corresponding to the dropped units from the previous and following layer respectively. As in standard FD, we only drop a fraction $\alpha$ of nodes per layer $i$, which produces the same model size for all clients.
For instance, in Fig.~\ref{fig:feddropoutexample} we have $N_{i}=N_{i+1}=5$ and $\alpha =2/5$ and therefore only $N_{i} \cdot N_{i+1} \cdot (1-\alpha )^2 = 25\cdot \frac{9}{25}= 9$ weights should be transmitted instead of $N_{i} \cdot N_{i+1} = 25$. 

The problem to be solved is to obtain a matrix $C_{i}$ per layer $i$ with the following properties:
\begin{itemize}
    \item each row of $C_{i}$ is a codeword $c^{k}_{i}\in \mathbf{R}^{N_{i}}$; 
    \item the Hamming weight (\ie the number of ones in the codeword) of each row in $C_{i}$ is equal to $N_{i} \cdot (1-\alpha)$; and
    \item the number of rows in $C_{i}$ is greater or equal than the number of clients per round M.
\end{itemize}
We consider 4 methods to compute $C_{i}$: (i) same random codeword for each client (baseline Federated Dropout), (ii)  different random codeword for each client (proposed contemporaneously by \cite{wen2021federateddropout}), (iii)  Gold sequences, and (iv) CWCs. While the first two are straightforward, the other two provide different levels of orthogonality between the dropped models. Since sub-models are trained independently and then aggregated, having partially non overlapping sub-models reduces the impact of heterogeneous updates.



In CFD we exploit the different masks per client. At the beginning of each training round, we compute one matrix $C_{i}$ for each layer i. Client k is assigned the k-th row from each $C_i$ and the correspondent weights are extracted from the global model. If such a matrix is burdensome to be computed or if it would be the same after the generation process, then the rows and columns of the matrix are shuffled instead. In that way, the excess codewords can be employed when the number of rows of $C_i$ is greater than M. 

Applying coded masks to neural networks is not straightforward, since it requires different approaches according to the kind of employed layers. For \textit{fully connected} layers, it is sufficient to apply dropout as in Fig.~\ref{fig:feddropoutexample}. For \textit{convolutional} layers, we experimentally observed that dropping entire filters rather than individual weights achieves better performance. \textit{Long Short-Term Memory} layers \cite{hochreiter1997long}, which are composed of multiple gates, require that the same mask is employed for each set of weights. 

After each sub-model has been trained, each weight in the global model is updated by averaging the contribution of each sub-model which contained that weight.




\subsubsection{Gold codes}
\label{sub:Gold}
Codewords orthogonality may be defined by means of cross-correlation. The correlation between two real binary sequences $u^1$ and $u^2$ of length $L_u$ is a function of the shift $l$:
\begin{equation}
\begin{split}
        & R(u_1,u_2,l) = \frac{1}{L_u}\sum_{j=1}^{L_u} u^1_j\cdot u^2_{(j+l) mod L_u}  \\
        & u^1_j,u^2_j \in \{-1,1\} \quad \forall j=1,...,L_u
\end{split}
\end{equation}


Gold codes are generated from two Linear Feedback Shift Registers (LFSR) with suitable feedback polynomials and initial conditions. The LFSRs produce two m-sequences, which are then circularly shifted and element-wise xored to produce all the sequences in the family. The size $n_{\text{LFSR}}$ of the LFSR determines the code length $2^{n_{\text{LFSR}}}-1$ , the number of codewords in the set $2^{n_{\text{LFSR}}}+1$, and the upper bounds for the maximum cross-correlation in the set:
\begin{equation}
\begin{split}
    & \max_{l}|{R({c}^{k_1}_i,{c}^{k_2}_i,l)}| = 2^{\lfloor{(n_{\text{LFSR}}+2)/2}\rfloor}+1\\
    & \forall k_{1},k_{2} \in \{1,...(2^{n_{\text{LFSR}}}+1)\},\quad k_{1}\neq k_{2}
\end{split}
\end{equation}
After computing the sequences, we concatenate them as row vectors in matrix $C_i$, which will then have  $2^{n_{\text{LFSR}}}+1$ rows and $2^{n_{\text{LFSR}}}$ columns.
Although Gold codes provide orthogonality, they have constraints on the size of $C_i$ and $\alpha$ value, which can only be  50\% since most Gold sequences are balanced (i.e., Hamming weight $2^{n_{\text{LFSR}}-1}$ ). Please refer to \cite{lfsrGoldSurvey} for details on constructions and properties of Gold codes and LFSR-generated sequences.
\subsubsection{Constant weight codes}
\label{sub:CWC}
Another metric is the Hamming distance between two codewords $u^1$ and $u^2$, which is the number of ones in $u^1 \oplus u^2$:
\begin{equation}
    d_h(u^1,u^2)= \sum_{j=1}^{L_u}{u^{1}_j \oplus u^{2}_j}
\end{equation}

We provide a method to create a matrix $C_i$ with size $M  \times N_i$ where, in order to improve orthogonality, the minimum Hamming distance between rows is maximized. CWC are a family of non-linear codes where each sequence has a fixed Hamming weight (\ie number of ones). CWC are flexible: they can provide sets of codewords with any cardinality, sequence length, and Hamming weight. We devise a variant of the algorithm from \cite{cwcheuristicconstruction} to generate $M$ codewords with length $N_i$ and weight $(1-\alpha)\cdot N_i$ by maximizing the minimum distance. Starting from the maximum possible minimum distance, we iteratively compute a codewords set of size $M$. If this is not feasible, we reduce the required minimum distance and repeat the process. The algorithm details can be found in \cite{cfdverardo}.




%% file: Evaluation.tex
\section{Evaluation}
\label{sec:evaluation}
We run our code in vanilla TensorFlow \cite{tensorflow} and Python3, since the major frameworks for FL do not support FD. In particular, although TensorFlow Federated \cite{TFF} allows FD with the same dropping masks, it does not allow broadcast of different models to the clients (which we require for CFD). The generation algorithm for CWC was implemented in Matlab. Training was performed on the EMNIST dataset \cite{cohen2017emnist} with convolutional model C \cite{modelC2015}. Preprocessing and full model description are provided in \cite{cfdverardo}. 
We use four kinds of codes for CFD with dropout fraction $\alpha=0.5$: random with same sub-model for each client (baseline FD), random with different sub-models, Gold and CWC. 

\vspace{.05in}
\noindent\textbf{The fast server learning rate tuning algorithm achieves consistent optimal $\eta$ values across many simulations}. We run 10 training sessions for each coded approach with target accuracy $\gamma^{*}={20}/{62}$ (which is 20 times the random accuracy) and showcase the results in Table~\ref{tab:fedadambestslrs}. Whereas for FAVG the selected $\eta$ is the same for all simulations  (except for the baseline FD), FedAdam produces higher variability. However, the differences between the different selected $\eta^*$ amounts to a maximum of 0.5 in log scale for each code. 

\begin{table}[!hbt]
\caption{Selected $\eta$ values for 10 simulations and different codes and aggregation algorithms}
\label{tab:fedadambestslrs}
\centering
\resizebox{\columnwidth}{!}{%
\begin{tabular}
{|l | r | r | r | r | r | r |}
\hline
                                      & \multicolumn{4}{c|}{\textbf{FedAdam}}                                                                                        & \multicolumn{2}{c|}{\textbf{FAVG}}                   \\ \hline
\textbf{Server Learning Rate (Log10)} & \multicolumn{1}{l|}{\textbf{-2.25}} & \multicolumn{1}{l|}{\textbf{-2}} & \multicolumn{1}{l|}{\textbf{-1.75}} & \textbf{-1.5} &  \multicolumn{1}{l|}{\textbf{0,25}}  & \textbf{0,5}   \\ \hline
\textbf{$\boldsymbol{\alpha}:$0.0 Fedadam No Dropout}    & 0\%            & 30\%        & 60\%  & 10\%          & 100\% & 0\%            \\ \hline
\textbf{$\boldsymbol{\alpha}:$0.5 Random Fedadam}        & 0\%            & 20\%        & 30\%          & \textbf{50\%} &  0\%            & \textbf{100\%} \\ \hline
\textbf{$\boldsymbol{\alpha}:$0.5 CWC FedAdam}           & 0\%            & 10\%        & 40\%           & \textbf{50\%} &  0\%           & \textbf{100\%} \\ \hline
\textbf{$\boldsymbol{\alpha}:$0.5 Gold FedAdam}          & 0\%            & 0\%         & \textbf{80\%}  & 20\%          &  0\%           & \textbf{100\%} \\ \hline
\textbf{$\boldsymbol{\alpha}:$0.5 Fedadam + Baseline FD} & \textbf{50\%}  & 50\%        & 0\%            & 0\%           &  \textbf{70\%}  & 30\%           \\ \hline

\end{tabular}
}
\end{table}

\begin{figure}[ht]
\centering
\subfigure[FedAdam]{
\includegraphics[width=0.80\columnwidth]{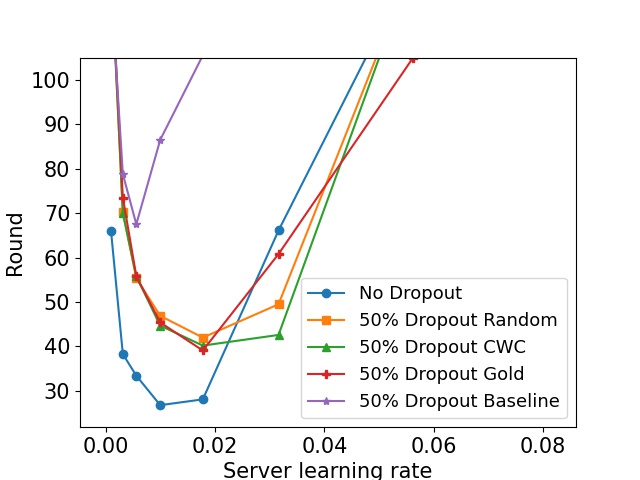}}
\subfigure[FAVG]{
\includegraphics[width=0.80\columnwidth]{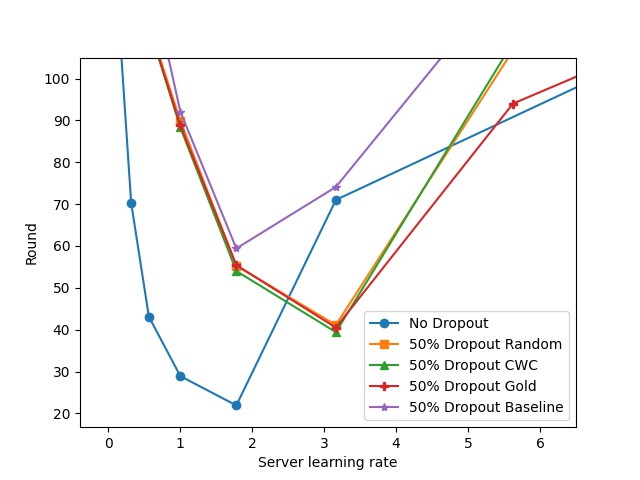}}
\caption{The dropout approaches require a higher $\eta$ in order to reach the minimal number of rounds for the target accuracy}
\label{fig:histograms}
\end{figure}

\noindent\textbf{The  optimal  server  learning  rate  in  a  traditional FL session is greater than that of an FD session, especially when a different sub-model per client is employed.} We experiment with 10 simulations with our tuning algorithm and keep track of the number of rounds to reach $\gamma^{*}$ for each experimented $\eta$.  Fig~\ref{fig:histograms} shows the average number of rounds for each $\eta$ for both FedAdam and FAVG and makes evident that the $\eta$ producing the minimum number of rounds is greater for the coded approaches compared to the no dropout. Moreover, although the no dropout case is still the fastest one, the coded approaches achieve up to $1.5\times$ speedup to reach $\gamma^{*}$ compared to the dropout baseline, thus improving convergence time. 

\noindent\textbf{Our tuning mechanism saves communication resources compared to running full FL sessions with different values of $\boldsymbol{\eta}$}. We compute the number of additional rounds as in Eq.~\ref{eq:commoverhead} and report the results in Table \ref{tab:additionalrounds}. In FedAdam\textsubscript{10},  $\gamma^{*}$ is 10 times the random accuracy (${10}/{62}$) instead of 20.
The overhead is directly dependent on the convergence speed of the model. Consequently, the no dropout case requires the least overhead and the baseline FD the most. Still, the additional number of rounds is much lower than running multiple full FL sessions. For both FedAdam and FAVG, our tuning algorithm tests 10 $\eta$ values. Therefore, running full sessions would require $500\cdot (10 - 1)$ additional rounds. We point out that decreasing the accuracy threshold $\gamma^{*}$ to ${10}/{62}$ notably reduces the additional number of rounds (FedAdam\textsubscript{10}). The best selected  $\eta$ is only sightly influenced by changing $\gamma^{*}$, thus demonstrating that tuning the threshold value is much easier than tuning $\eta$ directly.

\begin{table}[H]
\caption{Average number of additional rounds for different codes}
\label{tab:additionalrounds}
\resizebox{\columnwidth}{!}{%
\begin{tabular}{|c|c|c|c|c|c|}
\hline
\textbf{}        & \textbf{No Drop} & \textbf{Rand} & \textbf{CWC} & \textbf{Gold} & \textbf{FD} \\ \hline
\textbf{FedAdam\textsubscript{20}} & 154.9            & 262.5         & 248.6        & 259.3         & 456.3       \\ \hline
\textbf{FAVG\textsubscript{20}}    & 166.4            & 393.0         & 383.1        & 387.9         & 479.3       \\ \hline
\textbf{FedAdam\textsubscript{10}} & 140.7            & 219.2         & 225.6        & 211.9         & 351.2       \\ \hline
\end{tabular}
}
\end{table}

\begin{figure*}[!t]
\centering
\subfigure[FedAdam]{
\includegraphics[width=.43\textwidth]{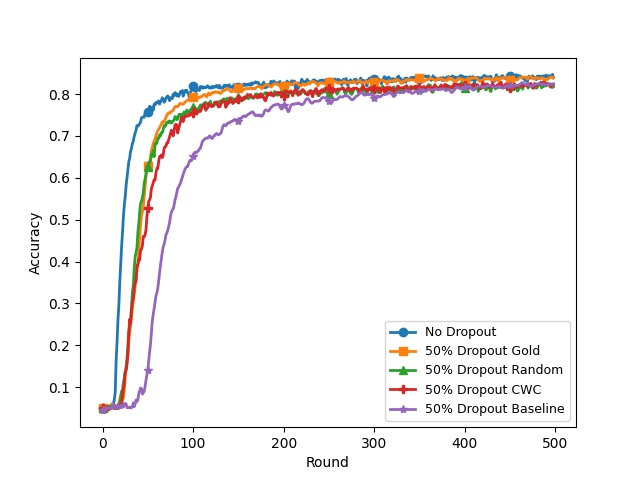}
}
\subfigure[FAVG]{
\includegraphics[width=.43\textwidth]{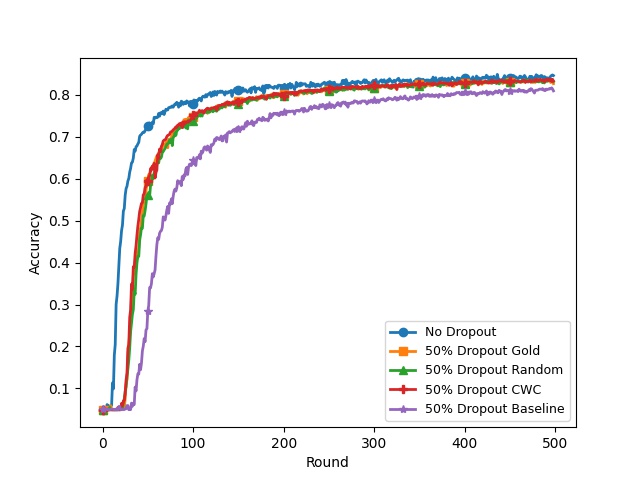}
}
\caption{Average test accuracy of FedAdam and FAVG  for 5 simulations with the selected $\eta^*$ from Table \ref{tab:fedadambestslrs}. Coded Federated Dropout with Gold codes and FedAdam improves convergence rate and final accuracy compared to random, CWC and baseline FD approaches.}
\label{fig:selectedlrssims}
\end{figure*}

\begin{figure*}[!t]
\centering
\subfigure[FedAdam]{
\includegraphics[width=.40\textwidth]{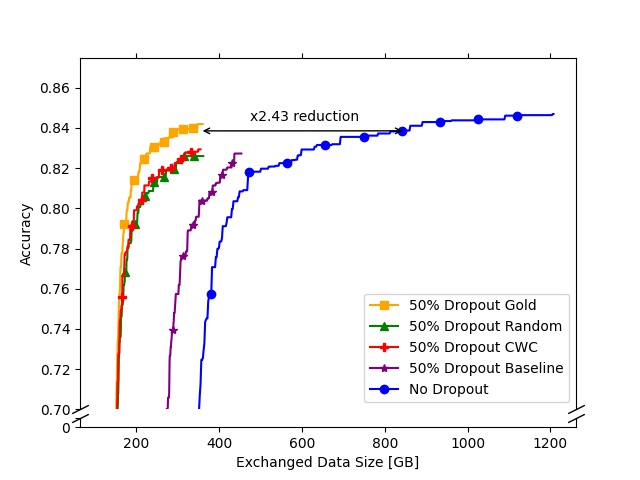}
}
\subfigure[FAVG]{
\includegraphics[width=.40\textwidth]{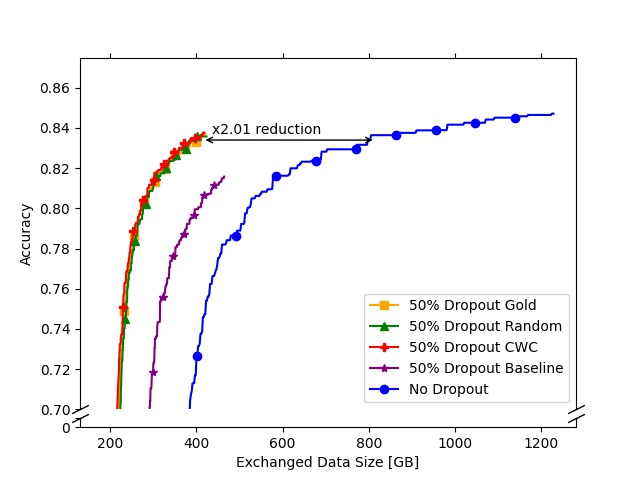}
}
\caption{Reachable test accuracy for FedAdam and FAVG given the amount of exchanged bytes in a FL session. Coded Federated Dropout with Gold codes and FedAdam reaches the same level of final accuracy as no dropout while reducing bandwidth usage by $2.43\times$.}
\label{fig:overheadreduction}
\end{figure*}

\noindent\textbf{Gold codes outperform other FD approaches for FedAdam and achieve 99.6\% of the final accuracy of the no dropout case while saving $>2\times$ bandwidth.} Fig.~\ref{fig:selectedlrssims} shows the average test accuracy of simulations run with the previously selected $\eta^*$ values. While the benefits of using Gold codes or CWC is negligible in terms of final accuracy compared to random for FAVG, FedAdam plus Gold codes produces higher convergence speed. Moreover, Gold FedAdam reaches 99.6 \% of the final accuracy of the no dropout case while saving almost $1-(1-\alpha )^2 = 75\%$ of the bandwidth per round. The best result is achieved by FAVG without dropout (84.1\% accuracy) when averaging over the last 100 rounds, while Gold codes achieve 83.8\% for FedAdam, which is the 99.6\%. Conversely, CWC does not perform well for FedAdam, achieving even worse performance than random codes. 
Regarding the overall bandwidth, we measure the size of the exchanged sub-models and compute the amount of gigabytes needed to reach a certain test accuracy. Fig.\ref{fig:overheadreduction} shows the reduction in overall bandwidth when CFD is employed with the selected $\eta^*$ values. Gold codes plus FedAdam reduces the bandwidth needed to reach the maximum test accuracy by $2.43\times$ compared to no dropout, while CFD plus FAVG by $2.01\times$. 

\noindent\textbf{Minimizing cross-correlation instead of maximizing minimal distance provides greater sub-model orthogonality}. Figures \ref{fig:selectedlrssims} and \ref{fig:overheadreduction} show that Gold codes always outperform CWC. Therefore, codes built by minimizing cross-correlation produce higher final accuracy and convergence rate than the ones obtained by maximizing the minimum distance. Nevertheless, optimizing any of the two metrics outperforms the baseline for federated dropout for both FedAdam and FAVG. 

%% file: Conclusion.tex
\section{Conclusion and future works}
We have presented a fast server learning rate tuning algorithm for Federated Dropout and shown considerable reduction on the number of rounds to assess the optimal $\eta^*$. Moreover, we have shown that convergence rate and final accuracy of models trained in a FL session are improved when using coding theory to carefully perform Federated Dropout. Specifically, CFD with Gold sequences paired with an optimization mechanism such as FedAdam can achieve up to the same accuracy of the no dropout case, with $2.43\times$ bandwidth savings. However, Gold codes have specific lengths and Hamming weights, so they are not flexible enough, while CWC does not improve performance compared to random dropout. Hence, future work will investigate further sequences from coding theory for FD.

%% file: Appendix.tex
\newpage

\section{Appendix}
\label{sec:appendix}
\begin{table*}[!t]
\centering 
\caption{Adopted notations and principal symbols}
\begin{tabular}{|l|l|l|}
\hline
\multicolumn{1}{|c|}{\bf{Symbol}} & \multicolumn{1}{|c|}{\bf{Description}} 
& 
\multicolumn{1}{|c|}{\bf{Value}} 
\\ \hline
\multicolumn{1}{|c|}{$N_i$}& Number of units in model layer i & See Tab. \ref{tab:emnist62model} \\ \hline
\multicolumn{1}{|c|}{$T$}& Number of total clients & \multicolumn{1}{|c|}{3400}\\ \hline
\multicolumn{1}{|c|}{$M$}&Number of clients per round & \multicolumn{1}{|c|}{35}\\ \hline
\multicolumn{1}{|c|}{$\alpha$}& Dropout fraction for FD  & \multicolumn{1}{|c|}{0.5}\\ \hline
\multicolumn{1}{|c|}{$\eta_l$}& Client learning rate  & \multicolumn{1}{|c|}{0.035}\\ \hline
\multicolumn{1}{|c|}{$E$}& Training epochs per client  & \multicolumn{1}{|c|}{1}\\ \hline
\multicolumn{1}{|c|}{$\eta$}& Server learning rate  & \multicolumn{1}{|c|}{Optimized}\\ \hline
\multicolumn{1}{|c|}{$\beta_1$}& Momentum parameter for FedAdam  & \multicolumn{1}{|c|}{0.90}\\ \hline
\multicolumn{1}{|c|}{$\beta_2$}& Momentum parameter for FedAdam  & \multicolumn{1}{|c|}{0.99} \\ \hline
\multicolumn{1}{|c|}{$\tau$}& Adaptivity degree for FedAdam  & \multicolumn{1}{|c|}{0.001} \\ \hline
\multicolumn{1}{|c|}{$w^{(t)}$} & Server weights at round t  & \multicolumn{1}{|c|}{-}         \\ \hline
\multicolumn{1}{|c|}{$w^{(t)}_{k}$} & Initial client k weights at round t   & \multicolumn{1}{|c|}{-}\\ \hline
\multicolumn{1}{|c|}{$\hat{w}^{(t)}_{k}$} & Final client k weights at round t   & \multicolumn{1}{|c|}{-}\\ \hline
\multicolumn{1}{|c|}{$D_k$} & Dataset for client k   & \multicolumn{1}{|c|}{-}\\ \hline
\multicolumn{1}{|c|}{$\{\xi\}^{(t)}_{k}$} &Set of batches for client k at round t   & \multicolumn{1}{|c|}{-}\\ \hline
\multicolumn{1}{|c|}{$L(.)$} & Loss function & \multicolumn{1}{|c|}{-}  \\ \hline
\multicolumn{1}{|c|}{$S(t)$} & Set of clients selected at round t  & \multicolumn{1}{|c|}{-} \\ \hline
\multicolumn{1}{|c|}{$c^{k}_{i}$} & Binary mask for client k and layer i  & \multicolumn{1}{|c|}{-} \\ \hline
\multicolumn{1}{|c|}{$c^{k}_{ij}$} & Component j of $c^{k}_{i}$  & \multicolumn{1}{|c|}{-} \\ \hline
\multicolumn{1}{|c|}{$C_{i}$} & Codes Matrix per layer i  & \multicolumn{1}{|c|}{-}\\ \hline
\multicolumn{1}{|c|}{$n_{LFSR}$} & Size of the LFSR & \multicolumn{1}{|c|}{See Tab.\ref{tab:preferredPolynomials}}\\ \hline
\multicolumn{1}{|c|}{$R(u_1,u_2,l)$} & Correlation between $u_1$ and $u_2$ for shift $l$ & \multicolumn{1}{|c|}{-}\\ \hline
\multicolumn{1}{|c|}{$d_h(u^1,u^2)$} & Hamming distance between $u_1$ and $u_2$ & \multicolumn{1}{|c|}{-}\\ \hline
\multicolumn{1}{|c|}{$w_h(u)$} & Hamming weight of $u$ & \multicolumn{1}{|c|}{-}\\ \hline
\multicolumn{1}{|c|}{$\gamma^{*}$} & Target accuracy & \multicolumn{1}{|c|}{$\frac{20}{62}$ or $\frac{10}{62}$}\\ \hline
\multicolumn{1}{|c|}{$\gamma_{k}^{t}$} & Training accuracy for client k at round t \\ \hline
\multicolumn{1}{|c|}{$\overline{\gamma}^{t}$} & Median training accuracy at round t & \multicolumn{1}{|c|}{-} \\ \hline
\multicolumn{1}{|c|}{$\overline{\gamma}$} & Average of $\overline{\gamma}^{t}$ at round t & \multicolumn{1}{|c|}{-}\\ \hline
\multicolumn{1}{|c|}{$q$} & Rounds number to compute $\overline{\gamma}$  & \multicolumn{1}{|c|}{-}\\ \hline
\multicolumn{1}{|c|}{$n_a$} & Number of adaptation steps  & \multicolumn{1}{|c|}{3}\\ \hline
\multicolumn{1}{|c|}{$\eta^{*}$} & Best server learning rate  & \multicolumn{1}{|c|}{See Tab.\ref{tab:fedadambestslrs}}\\ \hline
\multicolumn{1}{|c|}{$\Delta\eta$} & Log distance between tentative $\eta$ values & \multicolumn{1}{|c|}{-} \\ \hline
\multicolumn{1}{|c|}{$\Delta\eta_0$} & Initial $\Delta\eta$  & \multicolumn{1}{|c|}{1} \\ \hline
\multicolumn{1}{|c|}{$\mathbb{H}$} &  Set of tentative $\eta$ values & \multicolumn{1}{|c|}{-} \\ \hline
\multicolumn{1}{|c|}{$r$} & Current round number & \multicolumn{1}{|c|}{-} \\ \hline
\multicolumn{1}{|c|}{$r^{*}$} & Best round number to reach $\gamma^{*}$ & \multicolumn{1}{|c|}{-} \\ \hline
\multicolumn{1}{|c|}{$r^{*}_{i}$} & Best round number to reach $\gamma^{*}$ at step i & \multicolumn{1}{|c|}{-} \\ \hline
\end{tabular}

\label{tab:Notation}
\end{table*}
\subsection{Notation and parameter values}
The main system parameters employed throughout the paper and the related notation are summarised in Table \ref{tab:Notation}.

\subsection{Employed polynomials for Gold codes}
\begin{figure}[!htb]
\centerline{\includegraphics[scale=0.45]{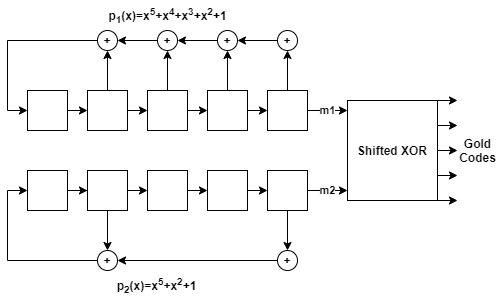}}
\caption{Gold code generation for preferred feedback polynomials $p_1(x)=x^5+x^4+x^3+x^2+1$ and $p_2(x)=x^5+x^2+1$. The output is a set of $2^5+1=32$ sequences of length $2^5-1=31$ }
\label{fig:lfsr}
\end{figure}

To generate Gold sequences, we employ the generation mechanism from Sec.~\ref{sub:Gold} with the preferred polynomials pairs listed in Table \ref{tab:preferredPolynomials}. The table also contains the correspondent length of the generated gold sequences. In order to have a suitable length for the model layers, we usually pad each resulting sequence with a 0 value after the longest run of zeros. Also note that Gold sequences with multiple of 4 degrees are not supported. Fig.~\ref{fig:lfsr} provides an example of the LFSR used to generate Gold codes of length 31.

\subsection{Generation Algorithm for Constant Weight Codes}
CWC are a non-linear class of codes.
Non-linearity implies that the generation of the sequences will be partially randomized, since a new codeword to be added in the set does not depend on the previously selected codewords. Hence, we follow the indications from \cite{cwcheuristicconstruction} and develop Algorithm \ref{alg:cwc}. Starting from a random codeword, we iteratively select new sequences with a fixed distance from the current set. The sequences are added in lexicographic order. If the final set is not complete (i.e., a set of $M$ sequences with weight $(1-\alpha)\cdot N_i$ and length $N_i$ does not exist for the given minimum distance), the required minimum distance is decremented and the selection procedure is performed again. The algorithm is described in \ref{alg:cwc}, where with an abuse of notation we identify $C_i$ as a set of codewords instead of a matrix.

\begin{algorithm}[!h]
\caption{Constant Weight Code Generation}
\label{alg:cwc}
\textbf{Input}: $N_i$, $M$, $\alpha$, $t_{max}$  \\
\textbf{Output}: $C_i$
\begin{algorithmic}[1] 
\STATE Let $t=0$.
\STATE $d_{min}=N_i$, $C_i=\{\}$ 
\STATE $F_i = \{ f \in \{0,1\}^{N_i} : w_h(f)= (1-\alpha)\cdot N_i \} $, f added in lexicographic order
\STATE Add random codeword from $F_i$ to $C_i$
\WHILE{$t < t_{max}$}
\STATE $F_i = \{ f \in F_i : d_h(f,c)\geq d_{min} \forall c \in C_i  \} $
\IF{$|F_i| \neq 0$}
\STATE Add first sequence from $F_i$ to $C_i$
\ELSE
\STATE $d_{min}=d_{min} - 2$,  $C_i=\{\}$ 
\STATE $F_i = \{ f \in {0,1}^N_i : w_h(f)= (1-\alpha)\cdot N_i \} $, f added in lexicographic order
\STATE Add random codeword from $F_i$ to $C_i$
\ENDIF
\IF{$|C_i| = M$}
\STATE \textbf{return} $C_i$
\ENDIF
\ENDWHILE
\STATE \textbf{return} $C_i$
\end{algorithmic}
\end{algorithm}

\begin{table*}[!htbp]
\centering
\caption{Preferred polynomial pairs for Gold codes generation}
\label{tab:preferredPolynomials}
\begin{tabular}{|r|r|l|l|}
\hline
\multicolumn{1}{|l|}{\textbf{Degree}} & \multicolumn{1}{l|}{\textbf{Sequence length}} & \textbf{Polynomial 1}                                                                          & \textbf{Polynomial 2}                                                                          \\ \hline
5               & 31                       & $1 + x^2 + x^5$                                                & $1 + x^2 + x^3 + x^4 + x^5$  \\ \hline
6               & 63                       & $1 + x^6 $                                                                      & $1 + x^1 + x^2 + x^5 + x^6$  \\ \hline
7               & 127                      & $1 + x^3 + x^7$                                                & $1 + x^1 + x^2 + x^3 + x^7$  \\ \hline
9               & 511                      & $1 + x^4 + x^9$                                                & $1 + x^3 + x^4 + x^6 + x^9$  \\ \hline
10              & 1023                     & $1 + x^3 + x^{10}$                                               & $1 + x^2 + x^3 + x^8 + x^{10}$ \\ \hline
11              & 2047                     & $1 + x^2 + x^5 + x^8 + x^{11}$ & $1 + x^2 + x^{11} $                                              \\ \hline
\end{tabular}
\end{table*}
\subsection{Datasets and models}
\subsubsection{EMNIST62}
We use the EMNIST carachter and digits recognition dataset provided by \cite{caldas2019leaf,TFF}, which includes grey scale images of size 28x28. We normalize each image in the $[0,1]$ interval and use a batch size of 10.
We provide a description of the employed model (with total number of parameters of 6,603,710) in Table~\ref{tab:emnist62model}. Each client optimize the sparse categorical crossentropy loss with the SGD optimizer. We train the model for 500 rounds.

\subsection{Practical notions about Federated Dropout}
As in \cite{caldasMCmahanReducingCommOverhead}, we perform federated dropout by dropping units (for dense layers) or filters (for convolutional layers) on each layer except the first and the last one.
\begin{table*}[!htbp]
\centering
\caption{EMNIST62 model summary}
\label{tab:emnist62model}
\begin{tabular}{|l|l|l|l|l|}
\hline
\textbf{Layer Type}     & \textbf{Output Shape} & \textbf{Param \#} & \textbf{Activation} & \textbf{Hyper-parameters}                                                                    \\ \hline
Conv2D                  & (-, 28, 28, 32)       & 832               & Relu                & \begin{tabular}[c]{@{}l@{}}Num filters: 32\\ Kernel size: (5,5)\\ Padding: Same\end{tabular} \\ \hline
MaxPooling2D            & (-,14,14,32)          & 0             & -                   & \begin{tabular}[c]{@{}l@{}}Pool size: (2,2)\\ Padding: Valid\end{tabular}                    \\ \hline
Conv2D                  & (-, 14, 14, 64)       & 51264             & Relu                & \begin{tabular}[c]{@{}l@{}}Num filters: 64\\ Kernel size: (5,5)\\ Padding: Same\end{tabular} \\ \hline
MaxPooling2D            & (-,7,7,64)            & 0                 & -                   & \begin{tabular}[c]{@{}l@{}}Pool size: (2,2)\\ Padding: Valid\end{tabular}                    \\ \hline
Flatten                 & (-, 3136)             & 0                 & -                   &                                                                                              \\ \hline
Dense                   & (-, 2048)             & 6424576           & Relu                & Num units: 2048                                                                              \\ \hline
Dense                   & (-, 62)               & 127038            & -                   & Num units: 62                                                                                \\ \hline

\end{tabular}
\end{table*}

%% file: ijcai22.bbl
\begin{thebibliography}{}

\bibitem[\protect\citeauthoryear{Abadi \bgroup \em et al.\egroup
  }{2016}]{tensorflow}
Martin Abadi, Paul Barham, Jianmin Chen, Zhifeng Chen, Andy Davis, Jeffrey
  Dean, Matthieu Devin, Sanjay Ghemawat, Geoffrey Irving, Michael Isard,
  Manjunath Kudlur, Josh Levenberg, Rajat Monga, Sherry Moore, Derek~G. Murray,
  Benoit Steiner, Paul Tucker, Vijay Vasudevan, Pete Warden, Martin Wicke, Yuan
  Yu, and Xiaoqiang Zheng.
\newblock Tensorflow: A system for large-scale machine learning.
\newblock In {\em 12th USENIX Symposium on Operating Systems Design and
  Implementation (OSDI 16)}, pages 265--283, 2016.

\bibitem[\protect\citeauthoryear{Alistarh \bgroup \em et al.\egroup
  }{2017}]{alistarh2017qsgd}
Dan Alistarh, Demjan Grubic, Jerry Li, Ryota Tomioka, and Milan Vojnovic.
\newblock Qsgd: Communication-efficient sgd via gradient quantization and
  encoding, 2017.

\bibitem[\protect\citeauthoryear{Alistarh \bgroup \em et al.\egroup
  }{2018}]{alistarh2018convergenceSParse}
Dan Alistarh, Torsten Hoefler, Mikael Johansson, Sarit Khirirat, Nikola
  Konstantinov, and Cédric Renggli.
\newblock The convergence of sparsified gradient methods, 2018.

\bibitem[\protect\citeauthoryear{Bouacida \bgroup \em et al.\egroup
  }{2020}]{bouacida2020adaptive}
Nader Bouacida, Jiahui Hou, Hui Zang, and Xin Liu.
\newblock Adaptive federated dropout: Improving communication efficiency and
  generalization for federated learning, 2020.

\bibitem[\protect\citeauthoryear{Brouwer \bgroup \em et al.\egroup
  }{1990}]{cwcnewtable}
A.E. Brouwer, J.B. Shearer, N.J.A. Sloane, and W.D. Smith.
\newblock A new table of constant weight codes.
\newblock {\em IEEE Transactions on Information Theory}, 36(6):1334--1380,
  1990.

\bibitem[\protect\citeauthoryear{Caldas \bgroup \em et al.\egroup
  }{2018}]{caldasMCmahanReducingCommOverhead}
Sebastian Caldas, Jakub Konecn{\'{y}}, H.~Brendan McMahan, and Ameet Talwalkar.
\newblock Expanding the reach of federated learning by reducing client resource
  requirements.
\newblock {\em CoRR}, abs/1812.07210, 2018.

\bibitem[\protect\citeauthoryear{Caldas \bgroup \em et al.\egroup
  }{2019}]{caldas2019leaf}
Sebastian Caldas, Sai Meher~Karthik Duddu, Peter Wu, Tian Li, Jakub Konecny,
  H.~Brendan McMahan, Virginia Smith, and Ameet Talwalkar.
\newblock Leaf: A benchmark for federated settings, 2019.

\bibitem[\protect\citeauthoryear{Chollet}{2015}]{kerasModels}
Fran\c{c}ois Chollet.
\newblock Keras.
\newblock \url{https://github.com/fchollet/keras}, 2015.

\bibitem[\protect\citeauthoryear{Cohen \bgroup \em et al.\egroup
  }{2017}]{cohen2017emnist}
Gregory Cohen, Saeed Afshar, Jonathan Tapson, and André van Schaik.
\newblock Emnist: an extension of mnist to handwritten letters, 2017.

\bibitem[\protect\citeauthoryear{{Gold}}{1967}]{goldcodes}
R.~{Gold}.
\newblock Optimal binary sequences for spread spectrum multiplexing (corresp.).
\newblock {\em IEEE Transactions on Information Theory}, 13(4):619--621, 1967.

\bibitem[\protect\citeauthoryear{Hochreiter and
  Schmidhuber}{1997}]{hochreiter1997long}
Sepp Hochreiter and J{\"u}rgen Schmidhuber.
\newblock Long short-term memory.
\newblock {\em Neural computation}, 9(8):1735--1780, 1997.

\bibitem[\protect\citeauthoryear{Karimireddy \bgroup \em et al.\egroup
  }{2021}]{karimireddy2021scaffold}
Sai~Praneeth Karimireddy, Satyen Kale, Mehryar Mohri, Sashank~J. Reddi,
  Sebastian~U. Stich, and Ananda~Theertha Suresh.
\newblock Scaffold: Stochastic controlled averaging for federated learning,
  2021.

\bibitem[\protect\citeauthoryear{Kasami}{1966}]{kasami1966weight}
Tadao Kasami.
\newblock Weight distribution formula for some class of cyclic codes.
\newblock {\em Coordinated Science Laboratory Report no. R-285}, 1966.

\bibitem[\protect\citeauthoryear{Lee}{1991}]{cdmamobile}
W.C.Y. Lee.
\newblock Overview of cellular cdma.
\newblock {\em IEEE Transactions on Vehicular Technology}, 40(2):291--302,
  1991.

\bibitem[\protect\citeauthoryear{Li \bgroup \em et al.\egroup
  }{2020}]{li2020convergence}
Xiang Li, Kaixuan Huang, Wenhao Yang, Shusen Wang, and Zhihua Zhang.
\newblock On the convergence of fedavg on non-iid data, 2020.

\bibitem[\protect\citeauthoryear{Li \bgroup \em et al.\egroup }{2021}]{ditto}
Tian Li, Shengyuan Hu, Ahmad Beirami, and Virginia Smith.
\newblock Ditto: Fair and robust federated learning through personalization.
\newblock In Marina Meila and Tong Zhang, editors, {\em Proceedings of the 38th
  International Conference on Machine Learning}, volume 139 of {\em Proceedings
  of Machine Learning Research}, pages 6357--6368. PMLR, 18--24 Jul 2021.

\bibitem[\protect\citeauthoryear{McMahan \bgroup \em et al.\egroup
  }{2017}]{mcmahan2017communicationefficient}
H.~Brendan McMahan, Eider Moore, Daniel Ramage, Seth Hampson, and
  Blaise~Agüera y~Arcas.
\newblock Communication-efficient learning of deep networks from decentralized
  data, 2017.

\bibitem[\protect\citeauthoryear{Montemanni and
  Smith}{2009}]{cwcheuristicconstruction}
Roberto Montemanni and Derek~H. Smith.
\newblock Heuristic algorithms for constructing binary constant weight codes.
\newblock {\em IEEE Transactions on Information Theory}, 55(10):4651--4656,
  2009.

\bibitem[\protect\citeauthoryear{Reddi \bgroup \em et al.\egroup
  }{2020}]{reddi2020adaptive}
Sashank Reddi, Zachary Charles, Manzil Zaheer, Zachary Garrett, Keith Rush,
  Jakub Konečný, Sanjiv Kumar, and H.~Brendan McMahan.
\newblock Adaptive federated optimization, 2020.

\bibitem[\protect\citeauthoryear{Sarwate and Pursley}{1980}]{lfsrGoldSurvey}
D.V. Sarwate and M.B. Pursley.
\newblock Crosscorrelation properties of pseudorandom and related sequences.
\newblock {\em Proceedings of the IEEE}, 68(5):593--619, 1980.

\bibitem[\protect\citeauthoryear{Springenberg \bgroup \em et al.\egroup
  }{2015}]{modelC2015}
Jost~Tobias Springenberg, Alexey Dosovitskiy, Thomas Brox, and Martin
  Riedmiller.
\newblock Striving for simplicity: The all convolutional net, 2015.

\bibitem[\protect\citeauthoryear{Srivastava \bgroup \em et al.\egroup
  }{2014}]{dropout}
Nitish Srivastava, Geoffrey Hinton, Alex Krizhevsky, Ilya Sutskever, and Ruslan
  Salakhutdinov.
\newblock Dropout: A simple way to prevent neural networks from overfitting.
\newblock {\em J. Mach. Learn. Res.}, 15(1):1929–1958, January 2014.

\bibitem[\protect\citeauthoryear{Taaghol \bgroup \em et al.\egroup
  }{1999}]{cdmasatellite}
P.~Taaghol, B.G. Evans, E.~Buracchini, G.~De~Gaudinaro, Joon~Ho Lee, and
  Chung~Gu Kang.
\newblock Satellite umts/imt2000 w-cdma air interfaces.
\newblock {\em IEEE Communications Magazine}, 37(9):116--126, 1999.

\bibitem[\protect\citeauthoryear{{tensorflow.org}}{2017}]{TFF}
{tensorflow.org}.
\newblock Tensorflow federated.
\newblock \url{https://www.tensorflow.org/federated}, 2017.

\bibitem[\protect\citeauthoryear{Verardo \bgroup \em et al.\egroup
  }{2022}]{cfdverardo}
Giacomo Verardo, Daniel Barreira, Marco Chiesa, and Dejan Kostic.
\newblock Fast server learning rate tuning for coded federated dropout, 2022.

\bibitem[\protect\citeauthoryear{Wen \bgroup \em et al.\egroup
  }{2021}]{wen2021federateddropout}
Dingzhu Wen, Ki-Jun Jeon, and Kaibin Huang.
\newblock Federated dropout -- a simple approach for enabling federated
  learning on resource constrained devices, 2021.

\bibitem[\protect\citeauthoryear{Zeiler}{2012}]{zeiler2012adadelta}
Matthew~D. Zeiler.
\newblock Adadelta: An adaptive learning rate method, 2012.

\end{thebibliography}
